\documentclass[sn-mathphys]{sn-jnl}

\usepackage{geometry}
\usepackage{amsthm,amsmath,amssymb,amsfonts}
 
\usepackage{graphicx}
\usepackage{multirow}
\usepackage{subcaption}
\usepackage{cleveref}
\usepackage{url}
\usepackage{booktabs}
\usepackage{natbib}

\jyear{2021}%

\begin{document}

\title[\textit{REDAffectiveLM}]{\textit{REDAffectiveLM}: Leveraging Affect Enriched Embedding and Transformer-based Neural Language Model for Readers' Emotion Detection}


\author*[1]{\fnm{Anoop} \sur{Kadan}}\email{anoopk\textunderscore dcs@uoc.ac.in}

\author[2]{\fnm{Deepak} \sur{P}}\email{deepaksp@acm.org}

\author[1]{\fnm{Manjary} \sur{P~Gangan}}\email{manjaryp\textunderscore dcs@uoc.ac.in}

\author[3]{\fnm{Savitha} \sur{Sam~Abraham}}\email{savitha.sam-abraham@oru.se}

\author[1]{\fnm{Lajish} \sur{V~L}}\email{lajish@uoc.ac.in}

\affil*[1]{
	\orgdiv{Department of Computer Science}, \orgname{University of Calicut},
	\orgaddress{
		\country{India}}}

\affil[2]{
	\orgdiv{School of Electronics, Electrical Engineering and Computer Science}, \orgname{Queen's University Belfast}, 
	\orgaddress{
		\state{Belfast}, \country{UK}}}

\affil[3]{
	\orgdiv{School of Science and Technology}, \orgname{Örebro University}, 
	\orgaddress{
		\country{Sweden}}}


\abstract{Technological advancements in web platforms allow people to express and share emotions towards textual write-ups written and shared by others. This brings about different interesting domains for analysis; emotion expressed by the writer and emotion elicited from the readers. In this paper, we propose a novel approach for Readers' Emotion Detection from short-text documents using a deep learning model called \textit{REDAffectiveLM}. Within state-of-the-art NLP tasks, it is well understood that utilizing context-specific representations from transformer-based pre-trained language models helps achieve improved performance. Within this affective computing task, we explore how incorporating affective information can further enhance performance. Towards this, we leverage context-specific and affect enriched representations by using a transformer-based pre-trained language model in tandem with affect enriched Bi-LSTM+Attention. For empirical evaluation, we procure a new dataset REN-20k, besides using RENh-4k and SemEval-2007. We evaluate the performance of our \textit{REDAffectiveLM} rigorously across these datasets, against a vast set of state-of-the-art baselines, where our model consistently outperforms baselines and obtains statistically significant results. Our results establish that utilizing affect enriched representation along with context-specific representation within a neural architecture can considerably enhance readers' emotion detection. Since the impact of affect enrichment specifically in readers' emotion detection isn't well explored, we conduct a detailed analysis over affect enriched Bi-LSTM+Attention using qualitative and quantitative model behavior evaluation techniques. We observe that compared to conventional semantic embedding, affect enriched embedding increases the ability of the network to effectively identify and assign weightage to the key terms responsible for readers' emotion detection to improve prediction.}




\maketitle

\section{Introduction}
\label{sec:introduction}

Social media advancements fueled by rapid developments in information technology have become an effective medium for expressing emotions across a wide variety of topics. New conventions that heavily use affective symbols like emojis and emotion reactions (e.g., emotion reactions in Facebook, Twitter, etc.) within text-based communication have enriched the density of emotion expression within social media. This deluge of social interactions provides two different perspectives for textual emotion detection research i.e., through \textit{`Writer Emotion'}, emotion expressed by the writer and \textit{`Readers' Emotion'}, emotion elicited from the readers. This poses an interesting \textit{dichotomy in textual emotion detection}, as the writer's intended emotions may not always be identical or in sync with the emotions generated for the readers. This makes readers' emotion detection an interesting arena for research. Considering the readers' perspective helps to infer emotion influence of the writer on readers', and also to understand other determinants of readers' emotions such as lexical word combinations or patterns in a document. These would enable novel applications such as \textit{emotion enabled information retrieval} for creation of emotion-aware search engines/recommendation systems \cite{chang2016linguistic},  \textit{emotion enriched article generation} using syntactic and semantic rules of language along with its emotional impact \cite{heaton2020language},  \textit{article auditing and writer influence forecasting} for automatically modulating emotionally sensitive contents, evaluating and regulating the {\it provocation potential of articles}, \textit{modeling of aesthetic emotion in poetry} \cite{haider2020po} and other tasks that can be conceptualized.

The works in literature that specifically address readers’ perspective of emotion detection \cite{bao2011mining,ye2012emotion,krebs2017social,anoop2022readers} are very few among the vast area of textual emotion detection. Methods for this may be categorized into three streams viz., lexicon based \cite{katz2007swat,bao2011mining}, classical machine learning \cite{bhowmick2009reader,ye2012emotion} and deep learning \cite{krebs2017social,anoop2022readers}. A major set of works are seen to be built over the backdrop of conventional semantic word embeddings \cite{krebs2017social,anoop2022readers}, which are powerful enough to identify similarities between words in near context; but a notable limitation due to the smaller window of neighboring words enables many times the contradictory affective words (emotion words) to share almost similar word representations (e.g. `good’ and `bad’) while learning these word embeddings \cite{socher2011semi}. This leads to degradation in performance among the affective computing related tasks such as sentiment analysis and emotion detection, and brings more suitable ways of word embeddings to encode affective information such as sentiment-specific \cite{tang2014learning} and affect enriched \cite{seyeditabari2019emotional,khosla2018aff2vec} embeddings. But even though an affective computing task, there has rarely been any work in text emotion detection that utilizes affect enriched word embedding \cite{chatterjee2019understanding} and none specific to readers' emotion detection, to our best knowledge. Textual emotion detection works in this context would be that of Chatterjee et al. \cite{chatterjee2019understanding}  proposing SS-BED, a sentiment specific word embedding, and Kratzwald et al. \cite{kratzwald2018deep} proposing sent2affect, a sentiment aided transfer learning from source network trained for sentiment analysis task to a target textual emotion detection network without direct affect enrichment in embedding; but both these works consider coarse-grained sentiment enrichment rather than affect enrichment required to suit the much fine-grained task of detecting diverse emotion classes. Even though the above mentioned representations/embeddings provide useful advancements, they are only capable of encoding the syntactic information and the word sense, but mostly miss to represent different meanings of the same word as a function of its context (e.g. the word `bank’ have different meanings in the context of words such as `river’ and `finance’). The recent transformer-based autoregressive and autoencoding pre-trained neural language models like BERT \cite{devlin2018bert}, GPT \cite{radford2018improving}, XLNet \cite{yang2019xlnet}, etc., have explored representing context specific, deeper and generic linguistic characteristics, thereby improving the performance, with the capability to fine-tune the architecture according to different NLP downstream tasks. These transformer-based language models are recently used in textual emotion detection \cite{adoma2020comparative}, even though not specifically in readers’ emotion detection, and are seen to obtain improved predictions. There are also works in textual emotion detection that combines transformer-based language model with graph convolutional network \cite{heaton2020language}, and Bi-LSTM learned from language-model \cite{adoma2020recognizing}; these works predominantly rely on context-specific representations learned from the transformer-based language models. 

\begin{figure}[!h]
	\centering
	\begin{subfigure}[b]{0.49\textwidth}
		\includegraphics[width=\linewidth,height=5cm]{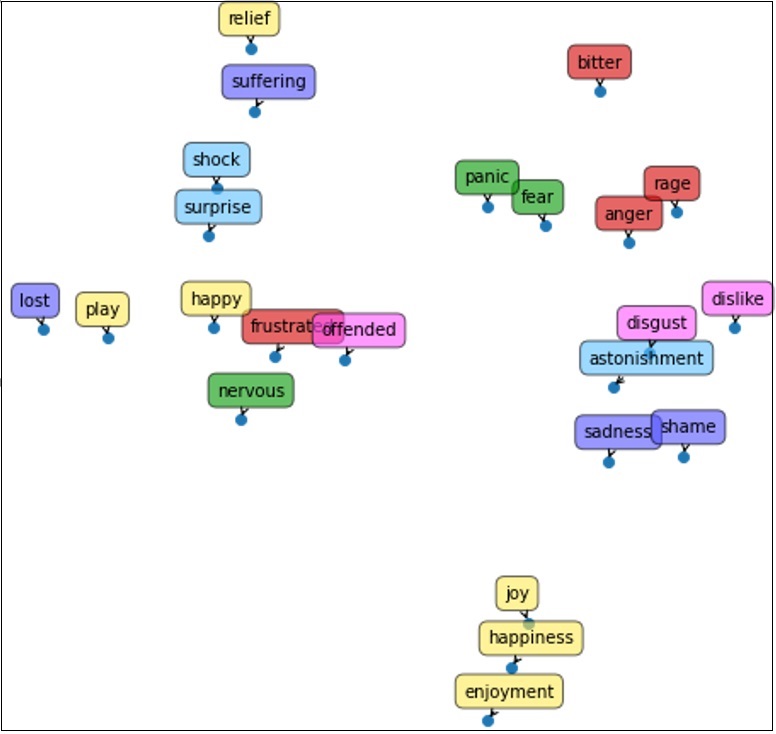}
		\caption{Conventional semantic embedding}
		\label{fig_glove}
	\end{subfigure}
	\begin{subfigure}[b]{0.49\textwidth}
		\includegraphics[width=\linewidth,height=5cm]{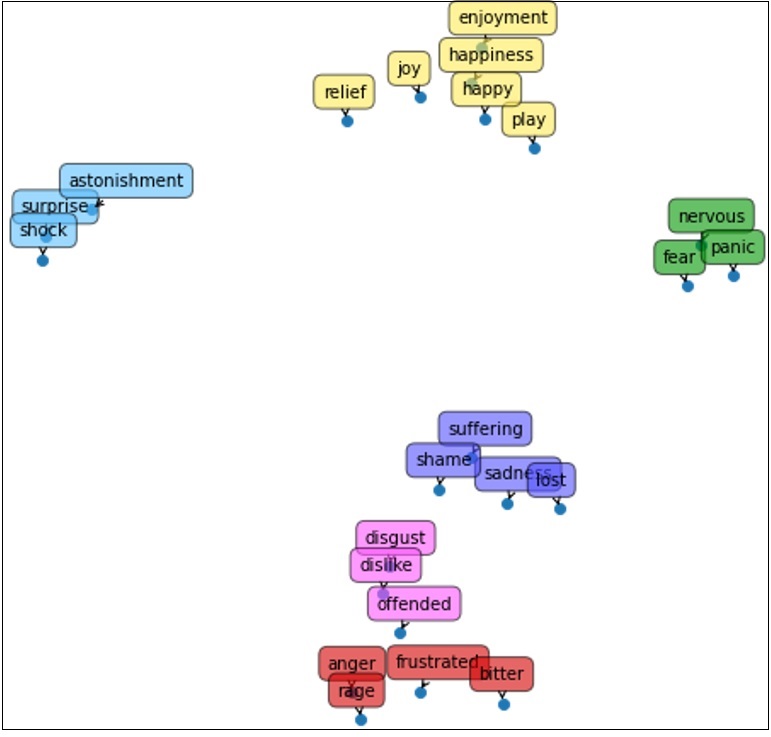}
		\caption{Affect enriched word embedding}
		\label{fig_emoglove}
	\end{subfigure}
	\caption{t-SNE visualization of few affective words related to basic emotions \colorbox[RGB]{220,103,103}{Anger}, \colorbox[RGB]{251,153,252}{Disgust}, \colorbox[RGB]{103,191,105}{Fear}, \colorbox{yellow!60}{Joy}, \colorbox{blue!40}{Sadness}, and \colorbox{cyan!35}{Surprise}}
	\label{fig_tsne}
\end{figure}

These context-specific representations from the pre-trained language models lack an explicit orientation towards representing affective information, something that is quite critical for affective computing tasks. We believe that utilizing affective information along with these context-specific representations would be highly beneficial for the task of readers’ emotion detection, as they are seen to produce better results when utilized in affective computing related tasks such as sentiment analysis, personality detection, etc., \cite{khosla2018aff2vec}. The t-SNE visualizations\footnote{\url{https://scikit-learn.org/stable/modules/generated/sklearn.manifold.TSNE.html}} of d-dimensional word representations of a few affective words, for conventional semantic embedding (GloVe \cite{pennington2014glove}) and affect enriched embedding (proposed in \cite{seyeditabari2019emotional}) shown in figure \ref{fig_tsne} demonstrates that compared to conventional semantic embedding, affect enrichment helps to cluster emotionally similar words into neighboring spaces. That is, affect enriched word embedding can encode affective information efficiently over conventional semantic embedding, which makes it more preferable for our task of readers' emotion detection than conventional semantic embedding. Therefore, we, to the best of our knowledge, for the first time attempt to leverage the utility of both the context-specific and affect enriched representations for the task of readers' emotion detection by proposing a deep learning based model \textit{REDAffectiveLM} built by fusing a transformer-based pre-trained language model with an affect enriched Bi-LSTM+Attention network. 

Our readers' emotion detection methodology is inspired from state-of-the-art research for natural language processing and affective computing that explores the combination of pre-trained language models with various networks to improve the overall model performance \cite{heaton2020language,adoma2020recognizing}. The choice of transformer-based pre-trained language model XLNet \cite{yang2019xlnet}, as we will detail, is motivated by its efficacy to combine the qualities of both autoregressive (e.g. GPT \cite{radford2018improving}) and autoencoding (e.g. BERT \cite{devlin2018bert}) pre-trained language models and produce improved performance over affective computing related tasks like sentiment analysis \cite{yang2019xlnet}. The choice of \textit{affect enriched Bi-LSTM+Attention} as the deep learning model is motivated by the pre-eminence of Bi-LSTM within related tasks and from the work proposed in \cite{khosla2018aff2vec} that demonstrates affect enrichment can improve performance of affective computing tasks. Bi-LSTM has the capability to learn long-term dependencies without keeping duplicate context representations and perform sequential modeling in both directions \cite{liang2016ac,jang2020bi}, and attention has the potential to enrich model performance \cite{kardakis2021examining} while also improving transparency of decision making and emerging as a prominent way of infusing interpretability within neural black box models \cite{sen2020human}. 

Our study is conducted over news documents that are short-text in nature, and we follow multi-target regression settings \cite{krebs2017social,tang2019hidden} that, beyond emotion classes, also provide information on emotion intensities, unlike the major category of single/multi-class or multi-label classification settings that only work on predicting emotion classes \cite{bhowmick2009reader,ye2012emotion,cabrera2020classifying}. Towards representing readers' emotions, similar to the works \cite{strapparava2007semeval,kratzwald2018deep,chatterjee2019understanding,anoop2022readers}, we utilize Paul Ekman’s discrete basic emotions \textit{anger}, \textit{disgust}, \textit{fear}, \textit{joy}, \textit{sadness}, and \textit{surprise} \cite{ekman1999basic}, the frequently discussed emotions  among the theorists in discrete emotion models, that also matches emotions provided in most online platforms for the readers to cast their emotions towards a post or news.
\\
The major contributions of this work are:
\begin{itemize} 
    \item We propose a novel deep learning approach for Readers' Emotion Detection called \textit{REDAffectiveLM} to predict readers' emotion profiles from short-text documents. This, in a novel direction, leverages both context-specific and affect enriched representations by fusing a transformer-based pre-trained neural language model and a Bi-LSTM+Attention network that utilizes affect enriched embedding.
    \item We evaluate the performance of our \textit{REDAffectiveLM} rigorously against a vast set of state-of-the-art baselines, where our method consistently outperforms baselines belonging to different categories of textual emotion detection, providing statistically significant improvements on fine-grained and coarse-grained evaluation measures. We also conduct a detailed analysis over the affect enriched Bi-LSTM+Attention network to understand the impact of affect enrichment specifically in readers' emotion detection using qualitative and quantitative behavior evaluation techniques.
    \item We procure a new Readers' Emotion News dataset REN-20k, with more than 20000 news documents and associated readers’ emotion profiles, to conduct our study. As our dataset also includes genre information of news documents, it can also be utilized for heterogeneous tasks such as document summarization and genre classification at various scales i.e., short-text and log-text. We shall contribute REN-20k at \url{https://dcs.uoc.ac.in/cida/resources/ren-20k.html} publicly as soon as this work is accepted for publication.
\end{itemize}

The rest of the paper is organized as, section \ref{sec:methodology} provides the detailed description of our proposed deep learning model for readers’ emotion detection followed by section \ref{sec:empirical_study} explaining the empirical study including details of the datasets, experimental settings, description of baselines and performance evaluation measures. The results and discussion in section \ref{sec:results_discussion} initially discuss the performance evaluation of our proposed model by comparing against the baselines, followed by the behavior analysis of affect enrichment in readers' emotion detection. Finally, section \ref{sec:conclusion} draws the conclusions.    

\section{Methodology} 
\label{sec:methodology}

This section presents our method for detecting Readers' Emotions from textual documents. We first discuss the problem settings followed by the architecture of our proposed model, \textit{REDAffectiveLM}. 

\subsection{Problem setting}
\label{sec:problem_setting}

We formulate our task of readers' emotions detection as a \textit{multi-target regression} problem. In other words, for each document, the model predicts readers' emotion profile i.e., intensities of the emotion classes \textit{anger}, \textit{fear}, \textit{joy}, \textit{sadness}, and \textit{surprise}. Each document $d$ consists of a sequence of $N$ words, $d=w_1,w_2,\ldots ,w_N$, where each word $w_i$ is taken from the vocabulary of $V$ unique words denoted by, $V=\left\{w_1,w_2,\ldots ,w_V\right\}$. The readers' emotion profile for each document, which forms the gold-standard labelled data for training, is formed from votes cast by multiple readers', which is normalized for $E$ distinct emotions, represented as, $ep_r(d)=\left\{e_1, e_2,\ldots ,e_E\right\}$, where, $e_i\in [0,1]$ and $\sum_{i=1}^E e_i=1$. Thus, the labelled corpus $D=\left\{(d_1,ep_r(d_1)),(d_2,ep_r(d_2)),\ldots ,(d_M,ep_r(d_M))\right\}$, represents $M$ documents along with their corresponding emotion profiles. We then follow a deep neural network based methodology to find the best fit mapping function $f: H\rightarrow ep_r(d)$, that predicts readers' emotion profile, $ep_r(d)$, for document vector $H$ of the document $d$.  

\subsection{Proposed Model}
\label{sec:proposed_model}

\begin{figure}[!h]
	\centering
	\includegraphics[scale=0.42]{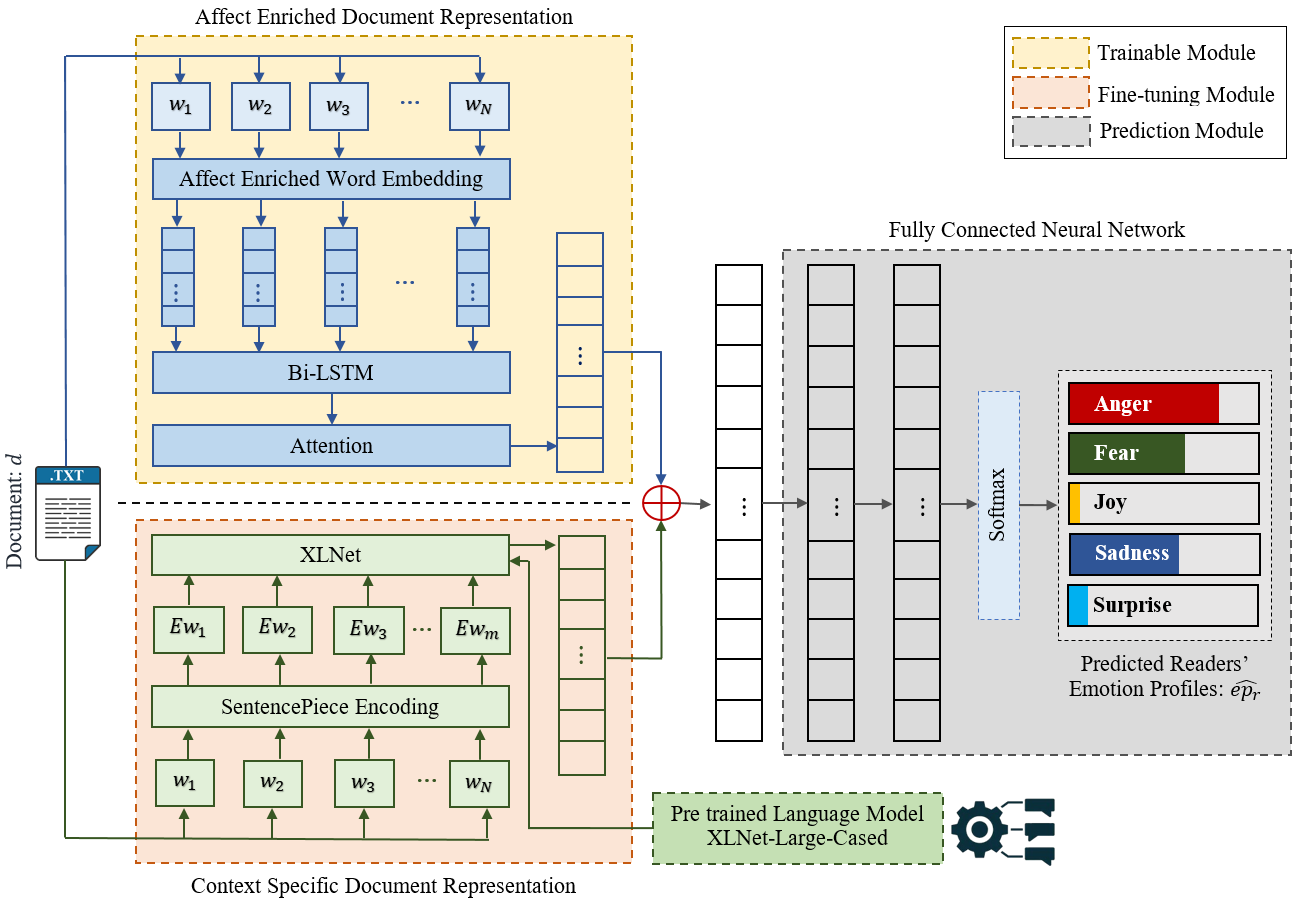}
	\caption{The proposed readers’ emotion detection system, \textit{REDAffectiveLM}}
	\label{fig:proposed_model}
\end{figure}

We propose a deep learning based readers’ emotion detection system, \textit{REDAffectiveLM} by parallely fusing two different networks, where the first emoBi-LSTM+Attention network is meant to produce affect enriched document representation and the second XLNet network for context-specific representation. We start by discussing the two networks in detail and later outline the complete architecture of our fused model, \textit{REDAffectiveLM}. An overall sketch of our proposed model is illustrated in figure \ref{fig:proposed_model}.

\subsubsection{emoBi-LSTM+Attention for Affect Enriched Document Representation} 
\label{sec:emobilstm}

In the emoBi-LSTM+Attention network, input documents are initially subject to Affect Enriched Word Embedding. 
To construct affect enriched word representations, denoted as emoGloVe, we utilize the state-of-the-art method using counter-fitting and emotional constraints\footnote{we have rewritten the code in \url{https://github.com/armintabari/Emotional-Embedding} from python 2.x to python 3.x to avoid compatibility issues with our implementations} proposed by Seyeditabari et al. \cite{seyeditabari2019emotional} over a pre-trained conventional semantic embedding GloVe \cite{pennington2014glove}. 
After generating affect enriched word representations, towards producing affect enriched document representations, we utilize Bi-LSTM \cite{schuster1997bidirectional}, a prominent RNN based architecture in combination with an Attention layer \cite{bahdanau2014neural}. The choice of Bi-LSTM network is motivated by its advantages such as the ability to learn long term dependencies \cite{liang2016ac} and to perform sequential modeling in both left to right ($\overrightarrow{forward}$) and right to left ($\overleftarrow{backward}$) directions which helps in producing excellent performance gains \cite{jang2020bi}. In addition, Attention on top of the Bi-LSTM network is observed to increase the overall model performance for the task of readers’ emotion detection \cite{anoop2022readers} and also the related task of sentiment analysis \cite{kardakis2021examining}. Attention’s mechanism of assigning corresponding weightages to words in the documents based on their relevance in emotion prediction also helps yet another objective of analyzing behavior of our network towards readers’ emotion detection. That is, in total, our choice of affect enriched word embedding and Bi-LSTM+Attention network, is based on the motivation that this combination should significantly contribute towards improving the overall model performance and moreover, allows to investigate the network behavior systematically to analyze the impact of affect enrichment in readers’ emotion detection. As we will see later (in section \ref{sec:affect_analysis}), we also analyze whether the attentions are influenced by emotion words and named entities, which are categories of words that we believe, hold much sway in determining affect. 

Bi-LSTM network is initially fed with the affect enriched word representations $\overrightarrow{w}_i$ of input document $d$, and it processes these sequential inputs in both forward and backward directions producing affect enriched contextual representation of the document as output vector. That is, a single layer $h_l$ in the Bi-LSTM network is defined as $[\overrightarrow{h}_l;\overleftarrow{h}_l]$, a concatenation of forward processing ($\overrightarrow{h}_l$) and backward processing ($\overleftarrow{h}_l$) hidden layers with parameters $\Theta_f$ and $\Theta_b$ respectively, denoted as,
\begin{align}
	\overrightarrow{h_l} &= LSTM(\overrightarrow{h}_{l-1},\overrightarrow{w}_i,\Theta_f)\\
	\overleftarrow{h_l} &= LSTM(\overleftarrow{h}_{l+1},\overrightarrow{w}_i,\Theta_b)
\end{align}
To build Attention on top of Bi-LSTM, we adopt the popular mechanism proposed in \cite{bahdanau2014neural}, where the final hidden layer of Bi-LSTM $h_n$ taken as the document summary vector $Z$ is passed to an alignment model, a feedforward network which is trained along with the model. With the learnable weight parameters $W_h, W_Z \in \mathbb{R}^{a\times b}$ and $v \in \mathbb{R}^a$, the alignment model generates a scalar value $u_i$, which on application of \textit{softmax} function delivers the set of word weightages $\alpha_i$ indicating the significance of each hidden state $h_i$ as, 
\begin{align}
	\text{i.e., } u_i &= v^{\top}\tanh(W_hh_i + W_ZZ)\\
	\alpha_i &= \frac{\exp(u_i)}{\sum_{j=1}^n \exp(u_j)}
\end{align}
Then, the final affect enriched document representation $H_1$ is computed as, 
\begin{align}
	H_1 = \begin{bmatrix}\alpha_1h_1 & \alpha_2h_2 & \alpha_3h_3 & \dots & \alpha_nh_n \end{bmatrix}
\end{align}

\subsubsection{XLNet for Context-specific Document Representation}
\label{sec:xlnet}

Transformer-based pre-trained language models are popular due to their efficacy in modeling linguistic relations and generating efficient context-specific document representations from various unlabelled text corpora; their effectiveness is evidenced by the promising results achieved for several downstream NLP tasks \cite{devlin2018bert,yang2019xlnet}. To learn such a document representation for our task, we adopt a popular transformer-based pre-trained language model, XLNet \cite{yang2019xlnet}, as the second network of our model. The choice of XLNet is motivated from its capability to enable bi-directional context representation through permutation of the factorization order, overcomes pretrain-finetune-discrepancy of autoencoding language models like BERT \cite{devlin2018bert}, and produces remarkable results for the very related affective computing task of sentiment analysis \cite{yang2019xlnet}. In the second network, initially, the text document, $d$ with a sequence of $N$ words, $d=w_1,w_2,\ldots ,w_N$,  is converted to encoded-word tokens, $EW = Ew_1, Ew_2, \dots , Ew_m$, using the popular \textit{SentencePiece} language-independent subword tokenization and detokenization module \cite{kudo2018sentencepiece}, where, $\text{\textbar} m \text{\textbar} \neq \text{\textbar} N \text{\textbar}$, and $Ew_i$ indicates encoded subword representation obtained by subdividing a single word into several subword units. The encoded data $EW$ is then fed to the pre-trained XLNet, which enables to fine-tune the architecture weights and hence to learn task-specific contextual document representations $H_2$ denoted as,
\begin{align}
	H_2 &= \text{XLNet}(EW)
\end{align}

\subsubsection{\textit{REDAffectiveLM}: The fused model for Readers' Emotion Detection}
\label{sec:redaffectivelm}

To build our Readers' Emotion Detection model, \textit{REDAffectiveLM}, that leverages the utility of Affect Enriched Document Representation and Context-specific Document Representation, we fuse the two networks, emoBi-LSTM+Attention and XLNet. In the fused model, the affect enriched document vector $H_1$ from emoBi-LSTM+Attention and context-specific document vector $H_2$ from XLNet are concatenated to form a single document vector $H$, defined as,
\begin{align}
	H = H_1 \oplus H_2
\end{align}
Finally, to predict readers' emotion profiles, we feed the concatenated document vector $H$ to a fully connected neural network module. Our neural network module that consists of a Multi-Layer Perceptron (MLP) having two fully connected dense hidden layers with 1224 neurons in each layer and an output layer with 5 neurons predicts normalized probability distribution of readers' emotion profiles $\widehat{ep_r(d)}$, given as,
\begin{align}
	\widehat{ep_r(d)} &= \text{\textit{softmax}}(\text{MLP}(H))
\end{align}
The learning process, computes and back propagates the loss between predicted emotion profile $\widehat{ep_r(d)}$ and labelled vector $ep_r(d)$. After model training, we empirically evaluate emotion prediction performance and later evaluate the impact of affect enrichment  qualitatively and quantitatively using document attention maps precipitated from the emoBi-LSTM+Attention network.

\section{Empirical Study}
\label{sec:empirical_study}

In this section we first describe the details of datasets used in our study, followed by the experimental settings, baselines and evaluation measures used for model performance analysis.

\subsection{Dataset}
\label{sec:dataset}

To conduct experiments we utilize three datasets, SemEval-2007 \cite{strapparava2007semeval}, RENh-4k \cite{anoop2022readers} and a newly curated \textit{Readers' Emotion News Dataset}, REN-20k. We detail these herewith. 

\subsubsection{SemEval-2007}
\label{sec:semeval_dataset}

SemEval-2007 \cite{strapparava2007semeval} is a popularly used short-text benchmark dataset collected from online news portals The New York Times, CNN, BBC and Google News. This is an annotated dataset with 1250 documents, where each document that comprises news headlines is annotated by six readers to obtain the scores of Anger, Disgust, Fear, Joy, Sadness and Surprise emotion classes.

\subsubsection{RENh-4k}
\label{sec:renh4k}

RENh-4k \cite{anoop2022readers} is a Readers’ Emotion News headlines dataset with 4000 news documents belonging to the year span 2015 to 2018 collected from the news portal Rappler\footnote{\url{https://www.rappler.com/}}. RENh-4k is short-text in nature, where each document comprises headlines and abstract of the news, and the corresponding emotion profiles of Afraid, Angry, Happy, Inspired, and Sad emotion classes are collected from the Mood Meter widget on the portal that records the percentage of votes cast by the readers for each emotion.

\subsubsection{REN-20k}
\label{sec:ren20k}

REN-20k is our newly curated Readers’ Emotion News dataset procured in a similar fashion of RENh-4k from the popular online news network Rappler, where we collect news articles manually, from the year span 2014 to 2019, by checking articles with high emotion votings in the Mood Meter widget of Rappler indicating high popularity and social reach of these articles. But it is an advanced version containing 20474 numbers of documents with corresponding readers' emotion profiles collected for diverse classes of emotions Afraid, Amused, Angry, Annoyed, Don't care, Happy, Inspired, and Sad. Also, each document consists of the whole news content including headlines, abstract and full-length news story excluding images and videos, making it a long-text dataset with average words per document as 527.84. With the help of genre information available in the portal and by manual annotations, we assign each document to a diverse set of genres, Business, Entertainment, Lifestyle, Sports, Technology and Others, unlike RENh-4k that considers only three genres, Health \& well-being, Social issues and Others. Since our work focuses on short-text documents, we choose only the headlines and abstract of news articles from each whole news document, to form the short-text version of REN-20k.

\subsection{Dataset Pre-processing}
\label{sec:dataset_preprocessing}

Since we utilize Paul Ekman's basic emotions \cite{ekman1999basic} \textit{anger}, \textit{disgust}, \textit{fear}, \textit{joy}, \textit{sadness}, and \textit{surprise} in our study, we perform an initial dataset preprocessing as followed in \cite{badaro2018emowordnet,anoop2022readers}, where emotion labels of RENh-4k and REN-20k taken from Rappler Mood Meter are mapped to basic emotions. That is, we map \textit{Angry}\textrightarrow\textit{Anger},  \textit{Sad}\textrightarrow\textit{Sadness}, \textit{Afraid}\textrightarrow\textit{Fear}, \textit{Happy}\textrightarrow\textit{Joy} and \textit{Inspired}\textrightarrow\textit{Surprise}, and discard other Mood Meter emotions such as \textit{Don't care}, \textit{Inspired}, \textit{Amused}, and \textit{Annoyed}. As followed by \cite{badaro2018emowordnet,anoop2022readers}, we exclude Paul Ekman's basic emotion \textit{Disgust} in our study as it does not have a matching emotion in the Rappler Mood Meter and for keeping a common set of five basic emotion labels across all the three datasets. We then perform a data normalization procedure where readers’ emotion profiles are represented as a distribution of the chosen five basic emotions, by adopting the technique in \cite{lei2014towards}. For better text representations, we perform data cleaning that removes frequently occurring noisy and unnecessary terms in the documents such as \textit{survey}, \textit{report}, \textit{(UPDATED)}, \textit{new-review} and \textit{Midday-wRa}, followed by other general set of pre-processing techniques such as text normalization and removal of punctuations and unknown symbols using NLTK toolkits\footnote{\url{https://www.nltk.org/}}. Table \ref{tab:dataset_statistics} shows the detailed dataset statistics after pre-processing where, to compute the number of annotations in RENh-4k and REN-20k we utilize the procedure in \cite{guerini2015deep}, as the number of annotators or readers’ are not known accurately, unlike six annotators explicitly mentioned in SemEval-2007.  

\begin{table}[h]
	\begin{center}
	\begin{minipage}{\textwidth}
		\caption{Dataset statistics after pre-processing}
		\label{tab:dataset_statistics}
		\begin{tabular*}{\textwidth}{@{\extracolsep{\fill}}llll@{\extracolsep{\fill}}}
			\toprule
			Statistics & REN-20k & RENh-4k & SemEval-2007\\ 
			\midrule
			Total number of words   & 10807161  & 124172  & 6364\\ 
			Number of unique words & 172243 & 13260  & 3286\\  
			Average words per document & 29.612 & 31.043 & 5.09\\  
			Average sentences per document  & 1.1826 & 1.1875 & 1.00\\  
			Number of annotations  & 2556654 & 242680 & 6 (\textit{annotators}) \vspace{3pt}\\ 
			\begin{tabular}[c]{@{}l@{}} 
				Mean percentage of votes\\for each emotion class\\ \\ \\ \\ 
			\end{tabular}  
			& \begin{tabular}[c]{@{}l@{}}
				Anger: 0.2253\\Fear: 0.0626\\Joy: 0.4222\\Sadness: 0.1441\\Surprise: 0.1459
			\end{tabular}  
			& \begin{tabular}[c]{@{}l@{}}
				Anger: 0.3388\\Fear: 0.1475\\Joy: 0.3137\\Sadness: 0.0781\\Surprise: 0.1218
			\end{tabular}
			& \begin{tabular}[c]{@{}l@{}}
				Anger: 0.1013\\Fear: 0.1639\\Joy: 0.2860\\Sadness: 0.2069\\Surprise: 0.2416
			\end{tabular} \vspace{3pt}\\  
			\begin{tabular}[c]{@{}l@{}}
				Number of articles associated\\with each emotion class\\ \\ \\ \\
			\end{tabular}
			&  \begin{tabular}[c]{@{}l@{}}
				Anger: 14419\\Fear: 8678\\Joy: 18104\\Sadness: 12841\\Surprise: 12749
			\end{tabular}
			&  \begin{tabular}[c]{@{}l@{}}
				Anger: 3068\\Fear: 1850\\Joy: 3267\\Sadness: 2489\\Surprise: 2312
			\end{tabular}
			& \begin{tabular}[c]{@{}l@{}}
				Anger: 652\\Fear: 820\\Joy: 786\\Sadness: 863\\Surprise: 1102
			\end{tabular}\\ \botrule
		\end{tabular*}
	\end{minipage}
	\end{center}
\end{table}

\subsection{Experimental Settings}

To conduct the experiments, each of the three datasets are split in the ratio 60:20:20 to form the corresponding train, validation, and test sets. In the emoBi-LSTM+Attention network, to develop affect enriched embedding emoGloVe, we consider embedding dimensions 300d and 100d, with various epochs 20, 50, 100, 150, 300, and 500, and later choose emoGloVe with dimension 100d and 20 epochs as a representative setting. The other hyperparameters in this network are the regularizer of Bi-LSTM set as \textit{l2(0.001)}, and \textit{dropout} between Bi-LSTM and Attention layer set as 0.5. To implement the XLNet architecture we utilize \textit{XLNet-Large-Cased} from the AI community Hugging Face\footnote{\url{https://huggingface.co/transformers/pretrained_models.html}}, where the hyperparameters are number of layers set as 24, hidden size as 1024, number of attention heads as 16, \textit{dropout} as 0.1, and altogether 360M trainable parameters fine-tune the network. In the fused model, affect enriched document vector $H_1$ from emoBi-LSTM+Attention network with dimension 200 and context-specific document vector $H_2$ from XLNet network with dimension 1024, on concatenation, forms a single final document vector $H$ with dimension 1224, which is then fed to a fully connected MLP. To build the MLP, we consider various number of layers having different combinations of neurons, such as, $\lbrace$1224\textrightarrow512\textrightarrow256\textrightarrow128\textrightarrow64\textrightarrow5$\rbrace$, $\lbrace$1224\textrightarrow1224\textrightarrow5$\rbrace$, $\lbrace$1224\textrightarrow5$\rbrace$, etc., and finally choose two hidden layers with 1224 neurons followed by the output layer with 5 neurons as a representative setting. The hyperparameters of our fused model \textit{REDAffectiveLM} are \textit{Adam} optimizer with learning rate 0.000015, Mean Squared Error (\textit{MSE}) as loss function, batch size as 64 and 200 epochs. \textit{REDAffectiveLM} consists of 363,762,235 number of total parameters, where 363,434,735 are trainable and 327,500 are non-trainable parameters.

\subsection{Baselines}
\label{sec:baselines}

To evaluate our \textit{REDAffectiveLM} model, we compare its performance using various measures (detailed in \ref{sec:performance_evaluation_measures}) against popular and state-of-the-art baselines belonging to lexicon based,  classical machine learning and deep learning categories (even though deep learning belongs under the umbrella of machine learning, we maintain deep learning baselines separately due to many notable contributions in the literature). Details of the baselines are as follows:
\begin{itemize} 
    \item Deep Learning Baselines: In our set of deep learning baselines we reproduce a vast number of works that follow a wide variety of methodology. That is our choice of deep learning baselines include a recent readers’ emotion detection work Readers’Affect \cite{anoop2022readers} that utilizes Bi-LSTM+Attention architecture, textual emotion detection works sent2affect \cite{kratzwald2018deep} that employs sentiment aided transfer learning and SS-BED \cite{chatterjee2019understanding} that utilize sentiment and semantic word embedding, a popular text classification architecture Kim's CNN \cite{kim2014convolutional}, naïve RNN architectures, GRU, LSTM and Bi-LSTM used as baselines in various textual emotion detection works \cite{kratzwald2018deep,chatterjee2019understanding,anoop2022readers}, and individual networks emoBi-LSTM+Attention and XLNet \cite{yang2019xlnet} used to construct our fused model (serving, implicitly as a form of ablation study).  
    \item Lexicon based Baselines: In this category of baselines we reproduce readers’ emotion detection systems, SWAT \cite{katz2007swat} popular among the top three systems of shared task `\textit{SemEval-2007 Task 14: Affective Text}' \cite{strapparava2007semeval} that utilize predefined sets of emotion words and Emotion Term Model \cite{bao2011mining} built over Naïve Bayes by incorporating emotion rating information and term independent assumption, and a promising textual emotion detection system Synesketch \cite{krcadinac2013synesketch} that utilizes word-level lexicon and an emoticon lexicon in hybrid with several heuristic rule sets. 
    \item Classical Machine Learning Baselines: In this category of baselines, we reproduce the textual emotion detection method proposed in \cite{ren2018emotion} that utilizes Word Mover's Distance (WMD) as a feature computed with the help of pre-trained word embeddings and provided to an SVM classifier (to implement this work, we instead use Support Vector Regression (SVR)\footnote{\url{https://scikit-learn.org/stable/modules/generated/sklearn.svm.SVR.html}} that suits our multi-target regression task). We also reproduce a wide variety of linguistic and affective features that are used in several textual emotion detection studies along with various multi-target regression models. The features considered are TF-IDF \cite{ren2018emotion,kratzwald2018deep}, N-Grams \cite{bandhakavi2017lexicon,chatterjee2019understanding}, General Purpose Emotion Lexicon Features \cite{bandhakavi2017lexicon} that includes Total Emotion Count (TEC), Total Emotion Intensity (TEI), Max Emotion Intensity (MEI), Graded Emotion Count (GEC), and Graded Emotion Intensity (GEI), extracted using the general purpose emotion lexicon DepecheMood++ \cite{araque2019depechemood++}, Sentiment word count feature \cite{bandhakavi2017lexicon,suharshala2018cross} computed using VADER \cite{gilbert2014vader}, Embedding Features \cite{chatterjee2019understanding,tang2014learning} that includes the semantic embeddings Word2Vec, GloVe and FastText, and Sentiment Specific Word Embedding, SSWE\textsubscript{u} proposed in \cite{tang2014learning}. For implementing the multi-target regression models, we adopt both problem transformation and algorithm adaptation techniques. In problem transformation approach we use Ridge Regression\footnote{\url{https://scikit-learn.org/stable/modules/generated/sklearn.multioutput.MultiOutputRegressor.html\#examples-using-sklearnmultioutput-multioutputregressor}}
    and in algorithm adaptation we use MLP\footnote{\url{https://keras.io/guides/sequential_model/}}. 
\end{itemize}
The hyper-parameters used to implement/reproduce the baselines GRU, LSTM, Bi-LSTM, Readers' Affect (Bi-LSTM+Attention) and emoBi-LSTM+Attention are, a single RNN stack, 100 neurons in a stack, pre-trained GloVe embedding with dimension 100, MSE loss function, Adam optimizer, softmax activation function in dense layer and 100 epochs. Except GRU for the other above-mentioned baselines regularizer is l2(0.001), dropout is 0.5, learning rate is 0.0005 and batch size is 128, whereas, for GRU, regularizer is l2(0.01), dropout is 0.25, learning rate is 0.005 and batch size is 64. For Kim’s CNN the number of filters are 100, filter sizes are 3, 4 and 5, dropout is 0.5, learning rate is 0.0005, and all other hyper-parameters are the same as GRU. For MLP, in the algorithm adaptation approach, we use a hidden layer with 128 neurons, ReLU activation, batch size of 64 and all other hyper-parameters are the same as LSTM. 

\subsection{Performance Evaluation Measures}
\label{sec:performance_evaluation_measures}

We choose various performance evaluation measures that are popularly used to evaluate textual emotion detection models \cite{bao2011mining,lei2014towards,liang2018universal,chatterjee2019understanding}. Accordingly,  for evaluating the performance of readers’ emotion detection we consider coarse-grained measures that look at correctness of our regression task by mapping predicted emotions to 0/1 classes, and fine-grained measures that look at nearness of predicted emotions to the ground truth at a finer granularity \cite{strapparava2008learning,anoop2022readers}. For coarse-grained evaluation, we use Acc@1 \cite{bao2011mining,lei2014towards,liang2018universal,anoop2022readers}, i.e., accuracy of the top first emotion prediction, representing micro-averaged F1 measure \cite{schutze2008introduction}. For fine-grained evaluations, we use correlation based measures AP\textsubscript{document} and  AP\textsubscript{emotion} \cite{liang2018universal,anoop2022readers} computing similarity of predicted emotion profiles with ground-truth, over emotions and documents respectively, and error/distance measures Root Mean Square Error (RMSE) \cite{chatterjee2019understanding,anoop2022readers} and Wasserstein Distance (WD) \cite{ghoshal2020estimating,anoop2022readers} computing the distance of predicted emotion profiles from ground-truth. 

\begin{itemize}
    \item Acc@1 of a corpus is an average of $Acc_d@1$ computed for all documents in the corpus. For the predicted emotion profile $X_d$ (shorthand for $\widehat{ep_r(d)}$) and ground-truth $Y_d$ (shorthand for ${ep_r(d)}$) of a document $d$, $Acc_d@1$ checks whether the  top-ranked emotion is the same for both prediction $\big(\text{i.e.}\arg \underset{i}{\max}\:{X_d[i]}\big)$ as well as ground-truth $\big(\text{i.e.}\arg \underset{i}{\max}\:{Y_d[i]}\big)$.  
        \begin{align}
            \text{i.e.,}\quad Acc_d@1 = \begin{cases}
                                        1 & \text{if, } \big(\arg\underset{i}{\max}\:{X_d[i]} = \arg \underset{i}{\max}\:{Y_d[i]}\big)\\ 
                                        0 & \text{else} 
                                    \end{cases} 
        \end{align}
    Since Acc@1 measures the accuracy, higher values are better.
    \item AP\textsubscript{document} of a corpus is the Average Pearson's correlation coefficient of all documents in the corpus obtained by averaging Pearson's correlation coefficient $P_d$ between prediction and ground-truth of each document $d$, over $\text{\textbar} E \text{\textbar}$ number of emotion classes.
        \begin{equation}
            P_d = \frac {\sum_{i=1}^{\text{\textbar} E\text{\textbar}}
                \big({X_d[i]}-{\overline{X}_d}\big)\big({Y_d[i]}-\overline{Y}_d\big)}
                {\big(\text{\textbar} E \text{\textbar}-1\big)\:\sigma_{X_d}\:\sigma_{Y_d}},  
                    \qquad P_d \in [-1,1]
        \end{equation}
        where, -1 and 1 indicate perfect negative and perfect positive correlations and $\overline{X}_d$, $\sigma_{X_d}$, $\overline{Y}_d$, $\sigma_{Y_d}$ indicate mean and standard deviation of predicted emotion profiles and ground-truth, respectively.
    \item AP\textsubscript{emotion} of a corpus is the Average Pearson's correlation coefficient of all the emotions obtained by averaging Pearson's correlation coefficient $P_e$ computed between prediction ($A$) and ground-truth ($B$) of each emotion category $e$ over $\text{\textbar} D \text{\textbar}$ number of documents.
            \begin{align}
                    P_e = \frac {\sum_{j=1}^{\text{\textbar} D \text{\textbar}}
                        \big(A_{j}-\overline{A}\big)\big(B_{j}-\overline{B}\big)}
                        {\big(\text{\textbar} D \text{\textbar}-1\big)\:\sigma_{A}\:\sigma_{B}},  
                            \qquad P_e \in [-1,1]
            \end{align}
    \item RMSE\textsubscript{D} of the corpus is an error metric computed by averaging RMSE of all documents. RMSE of a document ${d}$ is given by,
            \begin{align}
                RMSE_{d} =  \sqrt{\sum_{i=1}^{\text{\textbar} E \text{\textbar}} 
                            \frac{\big({X_d[i]} - {Y_d[i]}\big)^2}
                            {\text{\textbar} E \text{\textbar}}}
            \end{align}
            Since RMSE\textsubscript{D} measures the deviation between prediction and ground-truth, lower values are better.
    \item WD\textsubscript{D} of a corpus is a distance metric obtained by averaging WD of all documents in the corpus. WD of a document $d$ is the infimum for any transport plane computed as,
    \begin{align}
        WD_d\big(X_d,Y_d\big) = \inf_{\gamma\sim\pi(X_d,Y_d)} {\mathbb{E}}_{(x,y)\sim\gamma} [\parallel x-y\parallel]
        \end{align}
    where, $\pi(X_d,Y_d)$ is the set of all possible joint probability distribution $\gamma(x,y)$ whose marginals are $X_d$ and $Y_d$, respectively. Lower values of WD\textsubscript{D} indicate good performance.
\end{itemize}

\section{Results and Discussions}
\label{sec:results_discussion}

In this section we present the results of our experimental evaluations. Initially, we present the \textit{performance analysis} of our proposed \textit{REDAffectiveLM} model by comparing against a vast set of baselines from across families of lexicon based, classical machine learning, and deep learning including the individual emoBi-LSTM+Attention and XLNet networks of our model (that implicitly serves as a form of ablation study) to understand the gains achieved by our proposed model. We then perform statistical significance tests between our model and the best baseline. Finally, we also conduct \textit{behavior analysis} of the emoBi-LSTM+Attention network through a set of qualitative and quantitative experiments to identify the impact of affect enrichment for the task of readers’ emotion detection. 

\subsection{Model Performance Evaluation}
\label{sec:model_performance}

Experimental results of the evaluation measures over the REN-20k dataset for our \textit{REDAffectiveLM} model and the entire set of baselines are illustrated in table \ref{tab:results_ren20k}. From the results we can observe that our \textit{REDAffectiveLM} model achieve significant  gains\footnote{Here, we use ``\textit{gain}’’ to denote increase in percentage points ($\uparrow$) for measures Acc@1, AP\textsubscript {document} and AP\textsubscript {emotion}, and decrease in percentage points ($\downarrow$) for RMSE\textsubscript{D} and WD\textsubscript{D}} of 9.42, 4.68, 5.97, 5.7 and 6.19 percentage points for the evaluation measures Acc@1, AP\textsubscript {document}, AP\textsubscript {emotion}, RMSE\textsubscript{D} and WD\textsubscript{D}, respectively, when compared to the individual networks XLNet and emoBi-LSTM+Attention that archives best results in the category deep learning baselines, and 20.42, 21.07, 32.51, 17.9, and 11.48 percentage points when compared best results achieved by SWAT and Emotion Term Model in the category lexicon based baselines. For the classical machine learning category, we only provide N-Grams results for $N=1$ (unigrams) since it gives best results, similar to the observation in \cite{bandhakavi2017lexicon}. When comparing with problem transformation baselines, our model achieves a gain of 19.55, 17.79, 27.25, 17.62, and 9 percentage points, and with algorithm adaptation baselines a gain of 17.28, 16.03, 17.68, 15.73, and 8.96  percentage points for the same set of measures, respectively.

\begin{table}[h]
	\begin{center}
	\begin{minipage}{\textwidth}
	\caption{Evaluation results over REN-20k (Best results among all the models, and within each baseline category are highlighted in boldface)}
	\label{tab:results_ren20k}
	\setlength{\tabcolsep}{1pt}
	\begin{tabular*}{\textwidth}{@{\extracolsep{\fill}}lccccc@{\extracolsep{\fill}}}
		\toprule%
		\multicolumn{1}{c}{Model} & Acc@1(\%)$\uparrow$ & AP\textsubscript{document}$\uparrow$ & AP\textsubscript{emotion}$\uparrow$ & RMSE\textsubscript{D}$\downarrow$ & WD\textsubscript{D}$\downarrow$ \\ 
		\midrule
		\textit{REDAffectiveLM (Our Method)} & \textbf{76.68} & \textbf{0.8737} & \textbf{0.6806} & \textbf{0.0438} & \textbf{0.0104} \\ 
		\hline
		\multicolumn{6}{c}{Deep learning baselines} \\ 
		\hline
		sent2affect \cite{kratzwald2018deep} & 49.99 & 0.5925 & 0.1589 & 0.1945 & 0.1177 \\
		SS-BED \cite{chatterjee2019understanding} & 53.46 & 0.7114 & 0.4951 & 0.2197 & 0.1170 \\
		Kim's CNN \cite{kim2014convolutional} & 51.77 & 0.6228 & 0.1669 & 0.2285 & 0.1300 \\
		GRU \cite{anoop2022readers} & 53.47 & 0.6416 & 0.2202 & 0.2253 & 0.1221 \\
		LSTM \cite{chatterjee2019understanding} & 53.50 & 0.6866 & 0.4673 & 0.2192 & 0.1176 \\
		Bi-LSTM \cite{kratzwald2018deep} & 54.48 & 0.7077 & 0.5139 & 0.2165 & 0.1148 \\
		Bi-LSTM+Attention \cite{anoop2022readers} & 63.62 & 0.7998 & 0.5901 & 0.1277 & 0.0801 \\
		emoBi-LSTM+Attention & 65.09 & 0.8101 & \textbf{0.6209} & 0.1034 & 0.0800 \\
		XLNet \cite{yang2019xlnet} & \textbf{67.26} & \textbf{0.8269} & 0.6016 & \textbf{0.1008} & \textbf{0.0723} \\
		\hline
		\multicolumn{6}{c}{Lexicon based baselines} \\ 
		\hline
		SWAT \cite{katz2007swat} & 54.40 & \textbf{0.6630} & \textbf{0.3555} & \textbf{0.2228} & \textbf{0.1252} \\
		Emotion Term Model \cite{bao2011mining} & \textbf{56.26} & 0.6141 & 0.0245 & 0.3031 & 0.1999 \\
		Synesketch \cite{krcadinac2013synesketch} & 42.01 & 0.3375 & 0.2538 & 0.2594 & 0.1652 \\
		\hline
		\multicolumn{6}{c}{Problem transformation baselines} \\ 
		\hline
		WMD \cite{ren2018emotion} & 47.98 & 0.2571 & 0.2015 & 0.2508 & 0.1299 \\
		TF-IDF \cite{kratzwald2018deep,ren2018emotion} & 51.60 & 0.6746 & 0.3366 & 0.2298 & 0.1226 \\
		N-Grams   \cite{bandhakavi2017lexicon,chatterjee2019understanding} ($N=1$) & 50.74 & 0.5884 & 0.2939 & 0.2662 & 0.1247 \\
		TEC \cite{bandhakavi2017lexicon} & 55.94 & 0.6703 & 0.3524 & 0.2732 & 0.1112 \\
		TEI \cite{bandhakavi2017lexicon} & \textbf{57.13} & \textbf{0.6958} & \textbf{0.4081} & \textbf{0.2200} & 0.1106 \\
		MEI \cite{bandhakavi2017lexicon} & 54.37 & 0.6589 & 0.2901 & 0.2285 & 0.1176 \\
		GEC ($\delta=0.25$) \cite{bandhakavi2017lexicon} & 53.91 & 0.6588 & 0.3032 & 0.2268 & \textbf{0.1004} \\
		GEI ($\delta=0.25$) \cite{bandhakavi2017lexicon} & 53.86 & 0.6585 & 0.2919 & 0.2260 & \textbf{0.1004} \\
		Sentiment word count \cite{bandhakavi2017lexicon,suharshala2018cross} & 53.99 & 0.6389 & 0.2276 & 0.2299 & 0.1233 \\
		SSWE \cite{tang2014learning} ($d=50$) & 50.76 & 0.6080 & 0.1968 & 0.2234 & 0.1278 \\
		GloVe \cite{chatterjee2019understanding} ($d=100$) & 50.71 & 0.5939 & 0.1509 & 0.2240 & 0.1212 \\ 
		\hline
		\multicolumn{6}{c}{Algorithm adaptation baselines} \\ 
		\hline
		TF-IDF \cite{kratzwald2018deep,ren2018emotion} & 52.30 & 0.6563 & 0.2849 & 0.2257 & 0.1160 \\
		N-Grams \cite{bandhakavi2017lexicon,chatterjee2019understanding} ($N=1$) & 53.33 & 0.6073 & 0.3431 & 0.2291 & 0.1212 \\
		TEC \cite{bandhakavi2017lexicon} & 52.72 & \textbf{0.7134} & \textbf{0.5038} & 0.2027 & 0.1196 \\
		TEI \cite{bandhakavi2017lexicon} & \textbf{59.40} & 0.6824 & 0.3451 & 0.2207 & \textbf{0.1000} \\
		MEI \cite{bandhakavi2017lexicon} & 50.79 & 0.6035 & 0.2416 & 0.2325 & 0.1267 \\
		GEC ($\delta=0.25$) \cite{bandhakavi2017lexicon} & 53.04 & 0.6612 & 0.2906 & 0.2253 & 0.1139 \\
		GEI ($\delta=0.25$) \cite{bandhakavi2017lexicon} & 53.91 & 0.6456 & 0.2599 & \textbf{0.2011} & 0.1234 \\
		Sentiment word count \cite{bandhakavi2017lexicon,suharshala2018cross} & 52.54 & 0.6150 & 0.2176 & 0.2304 & 0.1225 \\
		SSWE \cite{tang2014learning} ($d=50$) & 50.79 & 0.5278 & 0.1051 & 0.3735 & 0.1309 \\
		GloVe \cite{chatterjee2019understanding} ($d=100$) & 51.06 & 0.5274 & 0.0613 & 0.3735 & 0.1309 \\ 		\botrule
		\end{tabular*}
	\end{minipage}
	\end{center}
\end{table}

Results of the models over RENh-4k illustrated in table \ref{results_renh4k} and SemEval-2007 illustrated in table \ref{results_semeval} show trends similar to REN-20k. In the results of RENh-4k, our model achieves significant gains of 7.38, 6.99, 6.22, 5.29 and 2.48 percentage points when compared to best results in deep learning category of baselines, 16.65, 18.35, 28.04, 13.56, and 8.47 percentage points compared to best lexicon based baseline results, 16.38, 17.85, 22.77, 12.04, and 5.55 percentage points compared to best problem transformation baseline results and 16, 16.64, 22.81, 12.01, and 5.82 compared to best algorithm adaptation baseline results for Acc@1, AP\textsubscript {document}, AP\textsubscript {emotion}, RMSE\textsubscript{D} and WD\textsubscript{D}, respectively. Similarly for SemEval-2007, the gains achieved by our model are 7.56, 6.13, 2.96, 6.90, and 3.75 percentage points compared to deep learning best results 17.56, 25.93, 25.21, 15.51, and 8.29 percentage points compared to best results in lexicon based baselines, 21.36, 23.35, 18.67, 11.26, and 6.1 percentage points compared to problem transformation best results and 17.36, 22.01, 15.09, 11.03, and 5.97 percentage points compared to algorithm adaptation best results for the same set of measures, respectively. The entire results over the three datasets thus consolidate that our \textit{REDAffectiveLM} model achieves best performance results when considering the top-ranked readers' emotion prediction (Acc@1) and overall readers' emotion profile prediction (AP\textsubscript {document} and AP\textsubscript {emotion}), and also obtains lower error/distance values (RMSE\textsubscript{D} and WD\textsubscript{D}). 

\begin{table}[h]
	\begin{center}
	\begin{minipage}{\textwidth}
\caption{Evaluation results over RENh-4k (Best results among all the models, and within each baseline category are highlighted in boldface)}
\label{results_renh4k}
\setlength{\tabcolsep}{1pt}
	\begin{tabular*}{\textwidth}{@{\extracolsep{\fill}}lccccc@{\extracolsep{\fill}}}
		\toprule%
		\multicolumn{1}{c}{Model} & Acc@1(\%)$\uparrow$ & AP\textsubscript{document}$\uparrow$ & AP\textsubscript{emotion}$\uparrow$ & RMSE\textsubscript{D}$\downarrow$ & WD\textsubscript{D}$\downarrow$ \\ 
		\midrule
\textit{REDAffectiveLM (Our Method)} & \textbf{60.75} & \textbf{0.7693} & \textbf{0.5809} & \textbf{0.1205} & \textbf{0.0761} \\ 
\hline
\multicolumn{6}{c}{Deep learning baselines} \\ \hline
sent2affect   \cite{kratzwald2018deep} & 36.00 & 0.4684 & 0.1047 & 0.2508 & 0.1458 \\
SS-BED   \cite{chatterjee2019understanding} & 45.62 & 0.5534 & 0.3609 & 0.2406 & 0.1424 \\
Kim's   CNN \cite{kim2014convolutional} & 40.00 & 0.4775 & 0.2084 & 0.2493 & 0.1585 \\
GRU \cite{anoop2022readers} & 38.75 & 0.4860 & 0.1765 & 0.2481 & 0.1443 \\
LSTM \cite{chatterjee2019understanding} & 40.13 & 0.5927 & 0.3402 & 0.2559 & 0.1472 \\
Bi-LSTM   \cite{kratzwald2018deep} & 45.00 & 0.6297 & 0.3415 & 0.2400 & 0.1465 \\
Bi-LSTM+Attention \cite{anoop2022readers}   & 50.50 & 0.6499 & 0.4054 & 0.2301 & 0.1220 \\
emoBi-LSTM+Attention & 51.98 & 0.6991 & \textbf{0.5187} & 0.1889 & 0.1141 \\
XLNet \cite{yang2019xlnet} & \textbf{53.37} & \textbf{0.6994} & 0.4975 & \textbf{0.1734} & \textbf{0.1009} \\
\hline
\multicolumn{6}{c}{Lexicon based baselines} \\ \hline
SWAT   \cite{katz2007swat} & 43.75 & \textbf{0.5858} & \textbf{0.3005} & \textbf{0.2561} & \textbf{0.1608} \\
Emotion   Term Model \cite{bao2011mining} & \textbf{44.10} & 0.5520 & 0.0102 & 0.3369 & 0.2000 \\
Synesketch   \cite{krcadinac2013synesketch} & 31.37 & 0.1394 & 0.2423 & 0.2936 & 0.1792 \\
\hline
\multicolumn{6}{c}{Problem transformation baselines} \\ \hline
WMD \cite{ren2018emotion} & 35.25 & 0.3593 & 0.0289 & 0.2869 & 0.1346 \\
TF-IDF          \cite{kratzwald2018deep,ren2018emotion} & \textbf{44.37} & 0.5007 & 0.3490 & 0.2440 & \textbf{0.1316} \\
N-Grams   \cite{bandhakavi2017lexicon,chatterjee2019understanding} ($N=1$) & 42.37 & 0.5067 & 0.3009 & 0.2662 & 0.1328 \\
TEC   \cite{bandhakavi2017lexicon} & 41.12 & 0.5686 & 0.3237 & 0.2410 & 0.1357 \\
TEI   \cite{bandhakavi2017lexicon} & 44.06 & \textbf{0.5908} & \textbf{0.3532} & \textbf{0.2409} & \textbf{0.1316} \\
MEI   \cite{bandhakavi2017lexicon} & 40.75 & 0.5394 & 0.2574 & 0.2442 & 0.1411 \\
GEC   ($\delta=0.25$)  \cite{bandhakavi2017lexicon} & 42.75 & 0.5676 & 0.3063 & 0.2410 & 0.1363 \\
GEI   ($\delta=0.25$)          \cite{bandhakavi2017lexicon} & 41.75 & 0.5602 & 0.2963 & 0.2417 & 0.1365 \\
Sentiment   word count \cite{bandhakavi2017lexicon,suharshala2018cross} & 39.25 & 0.4883 & 0.1443 & 0.2492 & 0.1386 \\
SSWE\textsubscript{u}   \cite{tang2014learning} ($d=50$) & 41.50 & 0.4969 & 0.1804 & 0.2483 & 0.1367 \\
GloVe   \cite{chatterjee2019understanding} ($d=100$) & 40.75 & 0.5108 & 0.2072 & 0.2474 & 0.1327 \\
\hline
\multicolumn{6}{c}{Algorithm adaptation baselines} \\ \hline
TF-IDF          \cite{kratzwald2018deep,ren2018emotion} & 39.62 & 0.4630 & 0.2870 & 0.2516 & 0.1489 \\
N-Grams   \cite{bandhakavi2017lexicon,chatterjee2019understanding} ($N=1$) & 42.75 & 0.4926 & 0.2796 & 0.2456 & 0.1505 \\
TEC   \cite{bandhakavi2017lexicon} & 41.37 & 0.5701 & 0.3298 & 0.2496 & 0.1356 \\
TEI   \cite{bandhakavi2017lexicon} & 42.87 & \textbf{0.6029} & \textbf{0.3528} & 0.2473 & \textbf{0.1343} \\
MEI   \cite{bandhakavi2017lexicon} & 40.12 & 0.4856 & 0.2279 & 0.2488 & 0.1466 \\
GEC   ($\delta=0.25$)  \cite{bandhakavi2017lexicon} & \textbf{44.75} & 0.5726 & 0.3190 & \textbf{0.2406} & 0.1359 \\
GEI   ($\delta=0.25$)          \cite{bandhakavi2017lexicon} & 41.37 & 0.5532 & 0.2934 & 0.2419 & 0.1378 \\
Sentiment   word count \cite{bandhakavi2017lexicon,suharshala2018cross} & 39.62 & 0.4846 & 0.1343 & 0.2491 & 0.1425 \\
SSWE\textsubscript{u}   \cite{tang2014learning} ($d=50$) & 35.62 & 0.3080 & 0.0207 & 0.4246 & 0.1376 \\
GloVe   \cite{chatterjee2019understanding} ($d=100$) & 35.37 & 0.2382 & 0.0920 & 0.4373 & 0.1376 \\
		\botrule
		\end{tabular*}
	\end{minipage}
	\end{center}
\end{table}

\begin{table}[h]
	\begin{center}
	\begin{minipage}{\textwidth}
\caption{Evaluation results over SemEval-2007 (Best results among all the models, and within each baseline category are highlighted in boldface)}
\label{results_semeval}
\setlength{\tabcolsep}{1pt}
	\begin{tabular*}{\textwidth}{@{\extracolsep{\fill}}lccccc@{\extracolsep{\fill}}}
		\toprule%
		\multicolumn{1}{c}{Model} & Acc@1(\%)$\uparrow$ & AP\textsubscript{document}$\uparrow$ & AP\textsubscript{emotion}$\uparrow$ & RMSE\textsubscript{D}$\downarrow$ & WD\textsubscript{D}$\downarrow$ \\ 
		\midrule
\textit{REDAffectiveLM (Our Method)} & \textbf{66.96} & \textbf{0.8235} & \textbf{0.6502} & \textbf{0.0902} & \textbf{0.0525} \\ 
\hline
\multicolumn{6}{c}{Deep learning baselines} \\ \hline
sent2affect \cite{kratzwald2018deep} & 37.20 & 0.3339 & 0.1075 & 0.2241 & 0.1428 \\
SS-BED   \cite{chatterjee2019understanding} & 50.40 & 0.6139 & 0.5098 & 0.1771 & 0.1090 \\
Kim's   CNN \cite{kim2014convolutional} & 47.20 & 0.5437 & 0.4451 & 0.1987 & 0.1200 \\
GRU \cite{anoop2022readers} & 46.00 & 0.5673 & 0.5003 & 0.2005 & 0.1098 \\
LSTM \cite{chatterjee2019understanding} & 49.20 & 0.6015 & 0.5248 & 0.1842 & 0.1089 \\
Bi-LSTM   \cite{kratzwald2018deep} & 49.89 & 0.6007 & 0.5059 & 0.1812 & 0.1074 \\
Bi-LSTM+Attention \cite{anoop2022readers} & 52.60 & 0.7140 & 0.5506 & 0.1700 & 0.0915 \\
emoBi-LSTM+Attention & 56.20 & 0.7565 & 0.5850 & \textbf{0.1592} & \textbf{0.0900} \\
XLNet \cite{yang2019xlnet} & \textbf{59.40} & \textbf{0.7622} & \textbf{0.6206} & 0.1739 & 0.0913 \\
\hline
\multicolumn{6}{c}{Lexicon based baselines} \\ \hline
SWAT \cite{katz2007swat} & 46.00 & 0.4945 & \textbf{0.3981} & \textbf{0.2453} & \textbf{0.1354} \\
Emotion Term Model \cite{bao2011mining} & \textbf{49.40} & \textbf{0.5642} & 0.0167 & 0.3031 & 0.1975 \\
Synesketch \cite{krcadinac2013synesketch} & 35.86 & 0.3705 & 0.3570 & 0.2470 & 0.1510 \\
\hline
\multicolumn{6}{c}{Problem transformation baselines} \\ \hline
WMD \cite{ren2018emotion} & 40.50 & 0.1447 & 0.0459 & 0.2430 & 0.1143 \\
TF-IDF          \cite{kratzwald2018deep,ren2018emotion} & \textbf{45.60} & 0.4954 & 0.4039 & 0.2080 & \textbf{0.1135} \\
N-Grams   \cite{bandhakavi2017lexicon,chatterjee2019understanding} ($N=1$) & 45.00 & 0.4992 & 0.3931 & 0.2089 & 0.1189 \\
TEC   \cite{bandhakavi2017lexicon} & 45.20 & 0.5451 & 0.4219 & \textbf{0.2028} & 0.1219 \\
TEI   \cite{bandhakavi2017lexicon} & \textbf{45.60} & \textbf{0.5900} & \textbf{0.4635} & 0.2985 & 0.1228 \\
MEI   \cite{bandhakavi2017lexicon} & \textbf{45.60} & 0.4884 & 0.4071 & 0.2051 & 0.1257 \\
GEC   ($\delta=0.25$)  \cite{bandhakavi2017lexicon} & 40.80 & 0.4643 & 0.3398 & 0.2113 & 0.1251 \\
GEI   ($\delta=0.25$)          \cite{bandhakavi2017lexicon} & 44.00 & 0.4416 & 0.3207 & 0.2136 & 0.1291 \\
Sentiment   word count \cite{bandhakavi2017lexicon,suharshala2018cross} & 39.04 & 0.5604 & 0.3820 & 0.2089 & 0.1208 \\
SSWE\textsubscript{u}   \cite{tang2014learning} ($d=50$) & 34.56 & 0.3130 & 0.1152 & 0.2300 & 0.1272 \\
GloVe   \cite{chatterjee2019understanding} ($d=100$) & 33.12 & 0.2605 & 0.1088 & 0.2378 & 0.1152 \\
\hline
\multicolumn{6}{c}{Algorithm adaptation baselines} \\ \hline
TF-IDF          \cite{kratzwald2018deep,ren2018emotion} & 46.40 & 0.4799 & 0.3941 & 0.2059 & 0.1206 \\
N-Grams   \cite{bandhakavi2017lexicon,chatterjee2019understanding} ($N=1$) & 46.80 & 0.5135 & 0.4140 & 0.2027 & 0.1171 \\
TEC   \cite{bandhakavi2017lexicon} & 46.40 & 0.5639 & 0.4270 & 0.2021 & 0.1204 \\
TEI   \cite{bandhakavi2017lexicon} & \textbf{49.60} & \textbf{0.6034} & \textbf{0.4993} & \textbf{0.2005} & \textbf{0.1122} \\
MEI   \cite{bandhakavi2017lexicon} & 46.40 & 0.4949 & 0.4103 & 0.2062 & 0.1306 \\
GEC   ($\delta=0.25$)  \cite{bandhakavi2017lexicon} & 46.00 & 0.4861 & 0.3622 & 0.2089 & 0.1229 \\
GEI   ($\delta=0.25$)          \cite{bandhakavi2017lexicon} & 46.70 & 0.4722 & 0.3531 & 0.2099 & 0.1248 \\
Sentiment   word count \cite{bandhakavi2017lexicon,suharshala2018cross} & 40.00 & 0.5732 & 0.3798 & 0.2023 & 0.1193 \\
SSWE\textsubscript{u}   \cite{tang2014learning} ($d=50$) & 40.80 & 0.2071 & 0.0595 & 0.4032 & 0.1641 \\
GloVe   \cite{chatterjee2019understanding} ($d=100$) & 42.40 & 0.2261 & 0.0777 & 0.4022 & 0.1643 \\
		\botrule
		\end{tabular*}
	\end{minipage}
	\end{center}
\end{table}

Across the three datasets, XLNet and emoBi-LSTM+Attention, individual networks of our model are the top two performing baselines in the deep learning category and even among the entire set of baselines belonging to the other categories. We believe this is because XLNet is a promising transformer based pre-trained language model that generates powerful contextual representations and, emoBi-LSTM+Attention enriches the conventional semantic representations with `affect’ that is evidently visible through the gains achieved by emoBi-LSTM+Attention (affect enriched) over Bi-LSTM+Attention (conventional) across the three datasets, for all the evaluation measures. But when comparing these individual networks with our model, the lowest among the gains achieved by our model, across the three datasets, are itself noteworthy. That is our model obtains a minimum gain of 7.38, 4.68, 2.96, 5.29 and 2.48 percentage points over XLNet and 8.77, 6.36, 5.97, 5.96, and 3.75 percentage points over emoBi-LSTM+Attention, for measures Acc@1, AP\textsubscript {document}, AP\textsubscript {emotion}, RMSE\textsubscript{D} and WD\textsubscript{D}, respectively. This indicates the promising nature of our \textit{REDAffectiveLM} model towards readers’ emotion detection, over these ablation or individual networks, leveraging both affect enriched document representation and contextual representation from transformer based pre-trained language model, effectively.

\begin{figure}[!h]
\centering
\captionsetup{type=figure}
\begin{subfigure}[b]{0.45\textwidth}
\centering
\includegraphics[width=\linewidth]{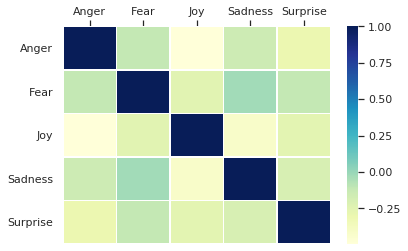}
\caption{REN-20k}
\label{ren20k}
\end{subfigure}
\begin{subfigure}[b]{0.45\textwidth}
\centering
\includegraphics[width=\linewidth]{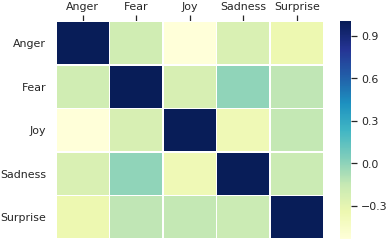}
\caption{RENh-4k}
\label{renh4k}
\end{subfigure}
\begin{subfigure}[b]{0.45\textwidth}
\centering
\includegraphics[width=\linewidth]{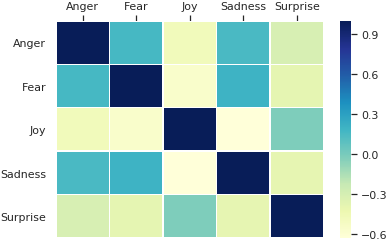}
\caption{SemEval-2007}
\label{semeval}
\end{subfigure}
\caption{Emotion profile correlations in the datasets}
\label{emotion_profile_correlation}
\end{figure}

Trends of evaluation results across the three datasets illustrate another point that the dataset SemEval-2007 obtains performance slightly better than RENh-4k even though it has comparably less data. This might be because SemEval-2007 being labeled by only six annotators is less complex in nature, whereas RENh-4k with 242680 and REN-20k with 2556654 minimum number of annotators make their ground truth emotion profiles complex in nature with several real-world contradictory readers' votings. Therefore, we analyze the complexity of datasets in terms of the number of reader annotations by computing Pearson's correlation \cite{tang2019hidden} between emotions and plot these correlations in figure \ref{emotion_profile_correlation}, where dark and light colors indicate high and low correlations respectively. SemEval-2007 shows many natural correlations, for e.g., high correlations that exist between \textit{anger}-\textit{fear}, \textit{anger}-\textit{sadness}, etc., but such correlations are less in RENh-4k and REN-20k. Similarly, in RENh-4k and REN-20k for contradictory emotion pairs, for e.g., \textit{joy}-\textit{fear} the correlations are comparatively higher than SemEval-2007. Such complex irregularities that emerge due to real-world scenario of differences in emotion expression among the readers might make it difficult for a model to generalize the learning process, thereby comparatively reducing performance for RENh-4k; but the vast amount of data in REN-20k, i.e., almost 5 times larger than RENh-4k, might be the reason for the model to overcome this difficulty in learning complex patterns, eventually obtaining noteworthy gains.

%
Besides looking into the performance gain obtained by our model over the baselines, we also analyze the statistical significance of our model performance with respect to Acc@1 and RMSE, the coarse-grained and fine-grained measures that ideally represent classification and regression task characteristics, respectively. We compute statistical significance between our \textit{REDAffectiveLM} model and the best performing baseline by conducting McNemar’s and Kolmogorov-Smirnov tests over Acc@1 and RMSE, respectively with conventional significance level (i.e., p-value 0.05). The p-values obtained for REN-20k, RENh-4k, and SemEval-2007 are 1.64E-5, 2.15E-3 and 5.07E-3 for Acc@1 and 1.80E-6, 3.47E-4 and 6.19E-4 for RMSE, respectively, indicating statistical significance of our model \textit{REDAffectiveLM} over the best baselines.

\subsection{Behavior Analysis of Affect Enrichment}
\label{sec:affect_analysis}

In this section, we analyze the impact of affect enrichment specifically for our task of readers' emotion detection, to verify its effectiveness over conventional semantic embedding. Besides the performance comparison of emoBi-LSTM+Attention (Bi-LSTM+Attention fed with affect enriched embedding) and Bi-LSTM+Attention (Bi-LSTM+Attention fed with conventional semantic embedding), in the above section \ref{sec:model_performance} by considering them as baselines in our empirical evaluation, here, we analyze the behavior of these networks. Our set of qualitative and quantitative behavior analysis compares the Attention Maps of emoBi-LSTM+Attention and Bi-LSTM+Attention precipitated from these attention based models along with readers' emotion profiles that highlight key terms with corresponding weightage based on their role in readers' emotion prediction (or decision making). That is, specifically, we look at the behavior of emoBi-LSTM+Attention network to understand whether the network has efficiently identified and assigned weightage to the key terms responsible for readers’ emotion detection (i.e., emotion words and named entities \cite{anoop2022readers}) to obtain significant performance gains in the predictions over Bi-LSTM+Attention.

\subsubsection{Qualitative Evaluation}
\label{sec:qualitative}

In qualitative behavior evaluation, we manually compare the key terms (emotion words and named entities) highlighted in the attention maps and their associated weightage, of both Bi-LSTM+Attention and emoBi-LSTM+Attention. Table \ref{table_attention_maps} shows pairs of attention maps for five sample documents, where in each pair, the first attention map is the one generated by Bi-LSTM+Attention and second by emoBi-LSTM+Attention, along with their associated ground-truth emotion profiles ($ep_r$) and predicted emotion profiles of both Bi-LSTM+Attention ($\widehat{ep}_{r}$) and emoBi-LSTM+Attention ($\widehat{ep}_{rEmo}$). In the attention maps, differing color intensities over the words represent weightage assigned to the words by the attention, i.e., dark red for high weightage and for lower weightage color intensities become lighter. 
		
\begin{table}[h]
	\begin{center}
	\begin{minipage}{\textwidth}
\caption{Sample attention maps}
\label{table_attention_maps}
\setlength{\tabcolsep}{2pt}
\begin{tabular*}{\textwidth}{@{\extracolsep{\fill}}lr@{\extracolsep{\fill}}}
\toprule%
\multicolumn{1}{c}{Document Attention Maps} 
& \multicolumn{1}{c} {\begin{tabular}[c]{@{}c@{}}Emotion profiles for\\{[}anger, fear, joy, sadness, surprise{]}\end{tabular}}     \\ \hline 
& $ep_{r}$ = {[}0.339, 0.122, 0.000, 0.245, 0.292{]}\vspace{2pt}\\ 
\includegraphics[width=0.5\textwidth, height=3mm]{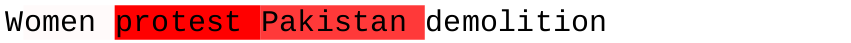}
    & $\widehat{ep}_{r}$ = {[}0.330, 0.210, 0.003, 0.280, 0.170{]} \\ 
\includegraphics[width=0.5\textwidth, height=3mm]{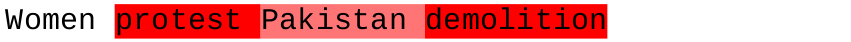} 
    & $\widehat{ep}_{rEmo}$ = {[}0.340, 0.102, 0.001, 0.290, 0.260{]} 
    \\ \hline 
& $ep_{r}$ = {[}0.551, 0.252, 0.045, 0.149, 0.000{]}\vspace{2pt}\\
\includegraphics[width=0.5\textwidth, height=3mm]{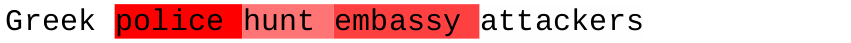}
    & $\widehat{ep}_{r}$ = {[}0.187, 0.277, 0.080, 0.301, 0.152{]} \\
\includegraphics[width=0.5\textwidth, height=3mm]{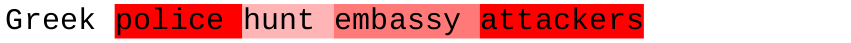} 
    & $\widehat{ep}_{rEmo}$ = {[}0.465, 0.272, 0.078, 0.103, 0.082{]} 
    \\ \hline
& $ep_{r}$ = {[}0.000, 0.000, 1.000, 0.000, 0.000{]}\vspace{2pt}\\
\includegraphics[width=0.5\textwidth, height=3mm]{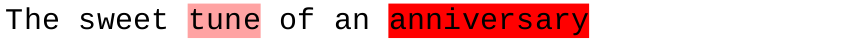}
    & $\widehat{ep}_{r}$ = {[}0.016, 0.026, 0.545, 0.247, 0.167{]} \\
\includegraphics[width=0.5\textwidth, height=3mm]{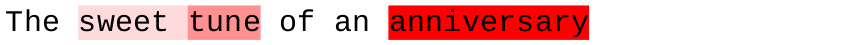} 
    & $\widehat{ep}_{rEmo}$ = {[}0.029, 0.039, 0.835, 0.064, 0.033{]} 
    \\ \hline
& $ep_{r}$ = {[}0.000, 0.495, 0.000, 0.221, 0.284{]}\vspace{2pt}\\
\begin{tabular}[c]{@{}l@{}} \includegraphics[width=0.5\textwidth, height=5mm]{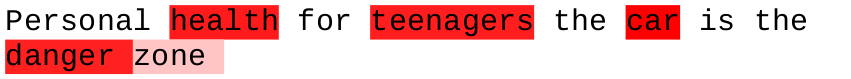}\\ \end{tabular}
    & $\widehat{ep}_{r}$ = {[}0.109, 0.229, 0.104, 0.349, 0.207{]} \\
\begin{tabular}[c]{@{}l@{}} \includegraphics[width=0.5\textwidth, height=5mm]{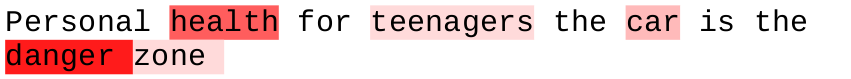}\\ \end{tabular}
    & $\widehat{ep}_{rEmo}$ = {[}0.056, 0.358, 0.080, 0.296, 0.210{]} 
    \\ \hline
& $ep_{r}$ = {[}0.000, 0.011, 0.915, 0.000, 0.074{]}\vspace{2pt}\\
\begin{tabular}[c]{@{}l@{}} \includegraphics[width=0.5\textwidth, height=10.5mm]{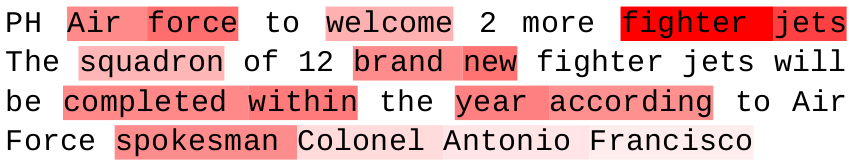} \end{tabular}
    & \begin{tabular}[c]{@{}l@{}} $\widehat{ep}_{r}$ = {[}0.093, 0.048, 0.606, 0.094, 0.159{]}\\ \\ \\ \end{tabular} \\
\begin{tabular}[c]{@{}l@{}} \includegraphics[width=0.5\textwidth, height=10.5mm]{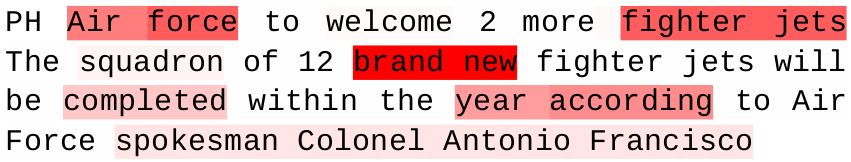} \end{tabular}
    & \begin{tabular}[c]{@{}l@{}} $\widehat{ep}_{rEmo}$ = {[}0.004, 0.039, 0.759, 0.004, 0.192{]} \\ \\ \\ \end{tabular} \\
    	\botrule
		\end{tabular*}
	\end{minipage}
	\end{center}
\end{table}

In the first pair, the attention map from Bi-LSTM+Attention significantly assigns weightage to an emotion word `protest’ and a named entity `Pakistan’. Whereas, the attention map from emoBi-LSTM+Attention shows improvements in the prediction, i.e., nearness of prediction to ground truth, especially visible in the case of emotions \textit{fear} and \textit{surprise} by assigning significant weightage to the emotion word `demolition’. In the second and third pair of attention maps, we can observe high improvements in prediction for emoBi-LSTM+Attention over Bi-LSTM+Attention, especially visible in the case of emotion \textit{anger} by identifying the emotion word `attackers’ in the second pair, and emotion \textit{joy} by identifying the emotion word `sweet’ in the third pair. Bi-LSTM+Attention, apart from failing to identify key terms (emotion words and named entities) such as `demolition’ in the first pair, `attackers’ in the second pair, `sweet’ in the third pair, etc., also are mostly seen to assign uniform weightage to the attention identified words. For example, in the fourth pair, the words `car’ and `teenager’ are given almost the same high intensity as the words `danger’ and `health’. But in the case of emoBi-LSTM+Attention, weightage for the words `car’ and ‘teenager’ are seen to be diminished than `danger’ and `health’. Similarly in the fifth pair, emoBi-LSTM+Attention assigns different weightage for the words `within’, `completed’, `year’, etc., whereas Bi-LSTM+Attention assigns almost similar weightage to these words. Hence the entire set of qualitative evaluations indicates that better than the Bi-LSTM+Attention that utilizes conventional semantic embedding, the affect enriched embedding based network emoBi-LSTM+Attention, can effectively identify and assign weightage to the key terms responsible for readers’ emotion detection thereby improving nearness of predictions to the ground-truth.

\subsubsection{Quantitative Evaluation}

Apart from the above mentioned qualitative behavior evaluation, we perform quantitative behavior analysis comparing capabilities of emoBi-LSTM+Attention and Bi-LSTM+Attention models in identifying key terms responsible for readers’ emotion detection. For quantitative analysis, we follow the approach similar to \cite{anoop2022readers} that checks the similarity between the External Attention Map representing the set of all emotion words and named entities in a document, and the Hybrid Attention Map representing the set of all emotion words and named entities in a document assigned with a weightage by attention mechanism; where high similarity between these attention maps indicate that the model efficiently utilizes key terms for decision making. External attention maps are binary maps created by highlighting only the emotion words and named entities in a document with a weightage of one and the rest with a zero weightage. Whereas, the hybrid attention map highlights only the emotion words and named entities in a document that have acquired non-zero attention weightage in both the model generated attention map and the external attention map, with weightage of the word copied from model generated maps; it can also be represented as binary maps by replacing the non-zero weightage with one. To create these attention maps, we use the popular emotion lexicons DepecheMood++ \cite{araque2019depechemood++} and EmoWordNet \cite{badaro2018emowordnet} and Named Entity Recognizer (NER) from spaCy\footnote{\url{https://spacy.io/}}. We generate external (EAM) and hybrid (HAM) attention maps for both emoBi-LSTM+Attention and Bi-LSTM+Attention models where for each model we contrast the extent of deviation between these attention maps using the similarity measures behavioral similarity, word similarity, and word probability \cite{anoop2022readers}, discussed below.

\begin{itemize} 
    \item Behavioral Similarity of a corpus $D$ is the average of pair-wise similarity between $\mbox{HAM}$ (taken as continuous) and $\mbox{EAM}$ for each document $d$ in the corpus.
        \begin{align}
            \text{BehSim\textsubscript{D}} = \frac{1}{D} \sum_{d=1}^{\text{\textbar} D \text{\textbar}} AUC(\mbox{HAM}_d, \mbox{EAM}_d)
        \end{align}
    where, $AUC$ is Area Under Curve that ranges from 0 (indicating negative similarity) to 1 (indicating perfect similarity) \cite{sen2020human}. 
    \item Word Similarity between is the average document cosine similarity\footnote{\url{https://deepai.org/machine-learning-glossary-and-terms/cosine-similarity}} between $\mbox{HAM}$ (taken as binary) and $\mbox{EAM}$.
    \begin{align}
        \text{WordSim\textsubscript{D}} = \frac{1}
        {\text{\textbar} D \text{\textbar} - \text{\textbar} D' \text{\textbar}}
        \sum_{d=1}^{\text{\textbar} D \text{\textbar} - \text{\textbar} D' \text{\textbar}} 
        \cos{(\mbox{HAM}_d, \mbox{EAM}_d)}
    \end{align}
    where, $\text{\textbar} D' \text{\textbar}$ is the total number of documents without any emotion words or named entities. 
    \item Word Probability of a corpus finds boolean intersection between binary HAM and EAM, averaged over the documents, to quantify how much among the total number of emotion words and named entities in the document are identified by attention.
    \begin{align}
        \text{WordProb\textsubscript{D}} = \frac{1}
        {\text{\textbar} D \text{\textbar} - \text{\textbar} D' \text{\textbar}}
        \sum_{d=1}^{\text{\textbar} D \text{\textbar} - \text{\textbar} D' \text{\textbar}} 
        \frac{\sum(\mbox{EAM}_d \cap \mbox{HAM}_d)}{\sum(\mbox{EAM}_d) + \lambda}
    \end{align}
    where,  $\lambda = 1$ for EAM = 0, and $\lambda = 0$ for EAM $\neq 0$. 
\end{itemize}

The results of quantitative analysis shown in table \ref{tab:quantitative} illustrate that for all the three datasets emoBi-LSTM+Attention obtains higher similarity scores between external and hybrid attention maps when compared to Bi-LSTM+Attention, for both the lexicons, which indicates that compared to Bi-LSTM+Attention model, emoBi-LSTM+Attention has improved ability to identify emotion words and named entities. Against the backdrop of \cite{anoop2022readers} that demonstrates emotion words and named entities are important for emotion detection, this validates emoBi-LSTM+Attention's improved suitability for emotion identification. Thus, the qualitative and quantitative behavior analysis on emoBi-LSTM+Attention together establishes that affect enrichment increases the ability of the model to effectively identify emotion words, and assign weightage to the key terms responsible for readers' emotion detection to improve prediction.

\begin{table}[h]
	\begin{center}
	\begin{minipage}{\textwidth}
\caption{Quantitative evaluation results}
\label{tab:quantitative}
\setlength{\tabcolsep}{2pt}
\begin{tabular*}{\textwidth}{@{\extracolsep{\fill}}lcccccc@{\extracolsep{\fill}}}
\toprule%
    \multicolumn{1}{c}{} 
        & \multicolumn{3}{c}{DepecheMood++} 
        & \multicolumn{3}{c}{EmoWordNet} 
        \\ \cmidrule(r){2-4} \cmidrule(l){5-7} 
    \multicolumn{1}{c}{Model} 
        & \begin{tabular}[c]{@{}c@{}}REN-\\20k\\ \end{tabular}  
        & \begin{tabular}[c]{@{}c@{}}RENh-\\4k\\ \end{tabular}  
        & \begin{tabular}[c]{@{}c@{}}SemEval-\\2007\end{tabular} 
        & \begin{tabular}[c]{@{}c@{}}REN-\\20k\\ \end{tabular} 
        & \begin{tabular}[c]{@{}c@{}}RENh-\\4k\\ \end{tabular}  
        & \begin{tabular}[c]{@{}c@{}}SemEval-\\2007\end{tabular}
    \\ \hline
    \multicolumn{7}{c}{Behavioral similarity scores ($\uparrow$)} 
    \\ \hline
    Bi-LSTM+Attention & 0.8829 & 0.7096 & 0.8092 & 0.8497 & 0.6988 & 0.8040\\
    emoBi-LSTM+Attention & 0.9537 & 0.8182 & 0.9001 & 0.9098 & 0.8104 & 0.8896
    \\ \hline
    \multicolumn{7}{c}{Word similarity scores ($\uparrow$)} 
    \\ \hline
    Bi-LSTM+Attention & 0.8296 & 0.6851 & 0.8203 & 0.8010 & 0.6606 & 0.7919\\
    emoBi-LSTM+Attention & 0.9603 & 0.8636 & 0.8821 & 0.8490 & 0.8128 & 0.8090
    \\ \hline
    \multicolumn{7}{c}{Word probability scores ($\uparrow$)} 
    \\ \hline
    Bi-LSTM+Attention & 0.9043 & 0.7648 & 0.8981 & 0.8901 & 0.7205 & 0.8624\\
    emoBi-LSTM+Attention & 0.9438 & 0.8071 & 0.8999 & 0.9413 & 0.7551 & 0.8873
    \\ 
        	\botrule
		\end{tabular*}
	\end{minipage}
	\end{center}
\end{table}

\section{Conclusion}
\label{sec:conclusion}

Context-specific representations from transformer-based pre-trained language models help textual emotion detection systems to achieve improved performance which, being an affective computing task, can be further enhanced by incorporating affective information. Inspired by this line of thought, in this paper, we proposed a novel deep learning model, \textit{REDAffectiveLM} that leverages context-specific and affect enriched representations by fusing a transformer-based pre-trained language model XLNet with, Bi-LSTM+Attention that utilizes affect enriched embedding, to predict readers’ emotion profiles from short-text documents. The performance of our proposed model was evaluated using coarse-grained and fine-grained measures, across three datasets, the benchmark SemEval-2007, RENh-4k and our newly procured REN-20k, where our model consistently outperformed a vast set of deep learning, lexicon based, and classical machine learning baselines in textual emotion detection and obtained statistically significant results. The evaluation results of our fused model \textit{REDAffectiveLM} when compared with the individual affect enriched Bi-LSTM+Attention and XLNet networks, obtained high gains in performance for all the evaluation measures, across all three datasets. This establishes that our \textit{REDAffectiveLM} model that utilizes highly efficient contextual representation from transformer-based pre-trained language model along with affect enriched document representation can significantly improve the performance of readers' emotion detection. We also performed a detailed qualitative and quantitative behavior analysis over affect enriched Bi-LSTM+Attention to study the impact of affect enrichment specifically in readers' emotion detection. We observed that compared to conventional semantic embedding, affect enrichment obtained higher performance and helped to increase the ability of the network to effectively identify and assign weightage to key terms (emotion words and named entities) responsible for readers' emotion detection. To aid future research, the datasets and other relevant materials, including the source code will be made publicly available at \url{https://dcs.uoc.ac.in/cida/projects/ac/redaffectivelm.html} and \url{https://github.com/anoopkdcs/REDAffectiveLM} soon as this work is accepted for publication. In the future, we are planning to explore the possibilities of developing affect enriched transformer-based language models. We are also planning to explore the applicability of affect enriched transformer-based language models in affective well-being tasks such as early detection of anxiety and depression from social networks. 

\backmatter

\bmhead{Acknowledgments}

The authors thankfully acknowledge the popular leading digital media company RAPPLER for the data source of news data along with associated emotions from their online portal that very relevantly helped to conduct this research. The authors thankfully acknowledge Arjun~K.~Sreedhar, Dheeraj~K., Sarath~Kumar~P.~S., and Vishnu~S.,  the postgraduate students of the Department of Computer Science, University of Calicut, who have been involved in dataset procurement. Manjary~P~Gangan was supported by the Women Scientist Scheme-A (WOS-A), Department of Science and Technology (DST) of the Government of India for Research in Basic/Applied Science under the Grant SR/WOS-A/PM-62/2018. 

\section*{Declarations}

\begin{itemize}
\item Funding: Not applicable
\item Conflict of interest/Competing interests: The authors declare that they have no competing interests
\item Ethics approval: Not applicable 
\item Consent to participate: Not applicable 
\item Consent for publication: The authors give the Publisher the permission to publish the work
\item Availability of data and materials: The dataset procured during the current study is available from the authors on reasonable request and also publicly available at \url{https://dcs.uoc.ac.in/cida/resources/ren-20k.html}
\item Code availability: Relevant materials, including the source code and datasets will be made publicly available at \url{https://dcs.uoc.ac.in/cida/projects/ac/redaffectivelm.html} and \url{https://github.com/anoopkdcs/REDAffectiveLM}
\item Authors' contributions: Anoop~Kadan, Deepak~P, and Lajish~V~L initiated the work. Anoop~Kadan and Deepak~P played key roles in conceptualization. Anoop~Kadan, Deepak~P, Manjary~P~Gangan and Savitha~Sam~Abraham designed the algorithm and experimental workflow. Anoop~Kadan and Manjary~P~Gangan obtained the datasets for the research, implemented and managed the coding. The rich experience of Deepak~P was instrumental in refining the work. The manuscript was collaboratively authored by Anoop~Kadan and Manjary~P~Gangan under the supervision of Deepak~P. All authors contributed to the editing and proofreading. All authors read and approved the final manuscript.
\end{itemize}









\bibliography{references_kais}


\end{document}